\documentclass[runningheads]{llncs}

\usepackage[mobile]{eccv}

\usepackage{eccvabbrv}

\usepackage{graphicx}
\usepackage{booktabs}

\usepackage[accsupp]{axessibility}  

\usepackage[breaklinks,colorlinks,bookmarksdepth=3,bookmarksopen=true,bookmarksopenlevel=2,bookmarksnumbered=true]{hyperref}

\usepackage{orcidlink}

\usepackage{enumerate}
\usepackage{wrapfig}
\usepackage{multirow}
\usepackage{bbm}
\newcommand{\parsection}[1]{\noindent\textbf{#1.}~}
\newcommand\dotfillin[1][3cm]{\makebox[#1]{\dotfill}}
\allowdisplaybreaks[4]

\begin{document}

\title{Motion-aware 3D Gaussian Splatting \\ for Efficient Dynamic Scene Reconstruction} 

\titlerunning{Motion-aware 3D Gaussian Splatting}


\author{
Zhiyang Guo\inst{1}\orcidlink{0000-0002-4278-1349} \and
Wengang Zhou\inst{1,2}$^{\star}$\orcidlink{0000-0003-1690-9836} \and
Li Li\inst{1}\orcidlink{0000-0002-7163-6263} \and\\
Min Wang\inst{2}\orcidlink{0000-0003-3048-6980} \and
Houqiang Li\inst{1,2}\thanks{Corresponding authors: Wengang Zhou and Houqiang Li.}\orcidlink{0000-0003-2188-3028}
}

\authorrunning{Z. Guo et al.}

\institute{CAS Key Laboratory of Technology in GIPAS, EEIS Department,\\
University of Science and Technology of China\\
 \and
Institute of Artificial Intelligence, Hefei Comprehensive National Science Center\\
\email{guozhiyang@mail.ustc.edu.cn, \{zhwg, lil1\}@ustc.edu.cn, wangmin@iai.ustc.edu.cn, lihq@ustc.edu.cn}}

\maketitle

\begin{abstract}
  3D Gaussian Splatting (3DGS) has become an emerging tool for dynamic scene reconstruction. However, existing methods focus mainly on extending static 3DGS into a time-variant representation, while overlooking the rich motion information carried by 2D observations, thus suffering from performance degradation and model redundancy.
  To address the above problem, we propose a novel motion-aware enhancement framework for dynamic scene reconstruction, which mines useful motion cues from optical flow to improve different paradigms of dynamic 3DGS.
  Specifically, we first establish a correspondence between 3D Gaussian movements and pixel-level flow.
  Then a novel flow augmentation method is introduced with additional insights into uncertainty and loss collaboration.
  Moreover, for the prevalent deformation-based paradigm that presents a harder optimization problem, a transient-aware deformation auxiliary module is proposed.
  We conduct extensive experiments on both multi-view and monocular scenes to verify the merits of our work.
  Compared with the baselines, our method shows significant superiority in both rendering quality and efficiency.

\end{abstract}

\section{Introduction}
\label{sec:intro}

In the domain of 3D vision, dynamic scene reconstruction is a crucial task with a wide range of applications, \eg, 3D animation and virtual/augmented reality. 
This task aims to model the three-dimensional structure and appearance of a scene that changes over time, thereby enabling novel view rendering at arbitrary timestamps.
While static scene modeling has witnessed significant progress in recent years, reconstructing dynamic scenes remains an intractable challenge due to the difficulties introduced by motion complexities, topological changes, and spatially or temporally sparse observations.

In the past few years, Neural Radiance Fields (NeRF)~\cite{nerf} has emerged as a remarkable implicit representation for 3D scenes. Researchers have developed various methods~\cite{dnerf,nerfies,hypernerf,nsff,du2021neural,attal2021torf,noguchi2021neural,hexplane} to model dynamic scenes with NeRF. Despite their impressive visual quality, the large time overhead stands as a non-negligible obstacle to their practical application.
Recently, a new method named 3D Gaussian Splatting (3DGS)~\cite{3dgs} has attracted substantial attention from the research community. 
3DGS gets rid of the expensive deep neural networks and ray-tracing rendering of NeRF-based methods by introducing the explicit 3D Gaussian representation and efficient point-based rasterization. 
As a strong competitor to NeRF, 3DGS achieves comparable performance in novel view synthesis while boosting the rendering speed to a real-time level.

It is then a straightforward but challenging task to extend the static 3DGS to a time-variant representation for dynamic content. Some pioneers~\cite{d3dgs,yang2023deformable,4dgs} have tried different strategies, \eg, iteration or deformation, to address this problem.
However, these 3DGS-based works typically focus on the design of dynamic modeling. They regard frames as discrete samples to fit time-dependent trajectories or deformations, while overlooking the rich motion cues underneath sequential 2D observations.
Since dynamic 3DGS involves explicit moving and deforming of Gaussians, it presents a tough under-constrained problem to use only images to supervise the reconstruction. 
The model is prone to local optimum where temporal consistency of Gaussians is not maintained as in physical world, especially for cases with insufficient viewpoints. 
This leads to visual overfitting, performance degradation, and redundant modeling in practice.

To address the above issues, we propose a novel motion-aware framework to enhance dynamic 3DGS by taking full advantage of optical flow prior. 
As a well-explored representation of pixel-level movement, optical flow can be efficiently predicted by pretrained networks, providing low-cost 2D motion prior for 3DGS. 
Instead of using plausible render-based supervision like previous practices in depth~\cite{dsnerf,depthnerf,chung2023depth} or segmentation~\cite{d3dgs}, we propose to establish a more robust and finer-grained cross-dimensional motion correspondence specially designed for flows.
In this way, Gaussian motions between frames can be aligned with 2D prior using our uncertainty-aware flow loss. 
Meanwhile, We offer dynamic awareness to existing regularization in neural rendering with the help of flow prior, thereby giving special attention to the motion parts during optimization.
Moreover, the prevalent deformation-based paradigm for dynamic 3DGS is susceptible to 3D motion ambiguities when relying solely on relative flow constraints. Therefore, we propose an additional deformation auxiliary module to inject transient motion information into Gaussian features and improve the dynamic modeling.
The overall framework is proved by extensive experiments to be an effective enhancing solution for multi-view/monocular scenes, which possesses efficient dynamic modeling capabilities with less redundancy.

Our main contributions can be summarized as follows:
\begin{itemize}
    \item To the best of our knowledge, we are the first to systematically explore the exploitation of flow prior in 3DGS-based dynamic scene reconstruction.
    \item We propose elaborate strategies, including uncertainty-aware flow augmentation and transient-aware deformation auxiliary, in order to develop an effective framework for enhancing different paradigms of dynamic 3DGS.
    \item Extensive experiments show that our method outperforms the baselines qualitatively and quantitatively in both multi-view and monocular scenes, enabling more accurate and efficient modeling of dynamic content.
\end{itemize}

\section{Related Works}

\subsection{Dynamic Neural Rendering with NeRF}

In computer vision, the necessity frequently arises for the reconstruction of dynamic 3D scenes. 
Since Neural Radiance Fields (NeRF)~\cite{nerf} signify a breakthrough in high-quality neural rendering, many efforts have been made to adapt NeRF to dynamic scenes thereafter.
Some works~\cite{dynerf,nsff} combine NeRF with time-conditioned latent codes to represent dynamic scenes.
Another category of methods~\cite{dnerf,nerfies,hypernerf,song2023nerfplayer} introduce an explicit deformation field to bend rays passing through varying targets into a canonical space, where a static NeRF is optimized.
With the development of efficient NeRF variants~\cite{eg3d,instantngp}, there are also approaches~\cite{kplanes,hexplane,tensor4d} attempting to factorize 4D spatio-temporal domain into 2D feature planes for compact model size. Although significantly accelerating the training process, they still cannot meet the practical need of rendering speed.

\subsection{Reconstruction with 3D Gaussian Splatting}

Recently, 3D Gaussian Splatting (3DGS)~\cite{3dgs} has emerged as a transformative technique in 3D representation and reconstruction. Characterized by the utilization of millions of explicit 3D Gaussians, this innovative method represents a significant departure from NeRF methodologies, promising not only real-time rendering capabilities but also unprecedented levels of control and editability.
Typically, 3DGS-based dynamic reconstruction is just unfolding in the research community.
D-3DGS~\cite{d3dgs} is proposed as the first attempt to extend 3DGS into a dynamic setup. Benefiting from the fast training and rendering, it employs an intuitive pipeline that iteratively performs frame-by-frame optimization. This paradigm is effective for multi-view dynamic scenes with long-term motions, but presents excessive memory consumption.
Inspired by the aforementioned methods for dynamic NeRF, 4D-GS~\cite{4dgs} and other concurrent works~\cite{yang2023deformable,katsumata2023efficient} introduce the deformation-based 3DGS that maintains only one set of canonical Gaussians and learns to deform them at each timestamp.
Notably, additional sensory information has been proved useful for regularization in 3DGS. By incorporating segmentation~\cite{d3dgs} or depth~\cite{chung2023depth} information into the optimization, the model tends to learn a better geometry.
However, the exploitation of flow-based motion priors in 3DGS still lacks exploration.
Render-based flow supervision as in NeRF~\cite{nsff} is a suboptimal and less robust expedient for 3DGS~\cite{katsumata2023efficient} due to the gap between pixel-level optical flow and blending 3D motion.
The explicit controllability of 3D Gaussians should enable a more flexible and finer-grained way to apply motion guidance.
In this work, we systematically explore the flow-guided motion awareness in 3DGS and provide an effective solution to this.

\section{Preliminary}

\subsection{3D Gaussian Splatting}

3DGS~\cite{3dgs} is an explicit representation using millions of 3D Gaussians to model a scene. Each Gaussian is characterized by a set of learnable parameters as follows:
\begin{enumerate}[1)]
    \item 3D center: $\mu \in \mathbb{R}^3$;
    \item 3D rotation represented by a quaternion: $\mathbf{q} \in \mathbb{R}^4$;
    \item 3D size (scaling factor): $\mathbf{s} \in \mathbb{R}^3$;
    \item view-dependent RGB color represented by spherical harmonics coefficients (degrees of freedom: $k$): $\mathbf{h} \in \mathbb{R}^{3(k+1)^2} \rightarrow \mathbf{c} \in \mathbb{R}^3$;
    \item opacity: $o \in [0,1]$.
\end{enumerate}
Each Gaussian can be regarded as softly occupying an area of space.
For a position $\mathbf{x} \in \mathbb{R}^3$ in the scene, the $i$-th Gaussian makes its contribution at that coordinate according to the standard Gaussian function weighted by its opacity:
\begin{equation}
    \varphi_{i}(\mathbf{x}) = o_i \exp(-\frac{1}{2}(\mathbf{x}-\mu_i)^T\mathbf{\Sigma}_{i}^T(\mathbf{x}-\mu_i)),
\label{eq:gaussian}
\end{equation}
where $\mathbf{\Sigma}_{i}\in \mathbb{R}^{3\times3}$ is the covariance matrix of the Gaussian calculated from $\mathbf{q}_i$ and $\mathbf{s}_i$.
Due to the characteristic of \cref{eq:gaussian}, each Gaussian has a long-distant (theoretically infinite) extent, so that gradients have global influence during optimization.
The differentiable rendering of 3DGS applies the splatting techniques~\cite{3dgs}. Given a view direction, $\mu$ and $\mathbf{\Sigma}$ of all Gaussians are projected to the 2D camera plane. Then the density $\alpha$ of any 3D point can be calculated via a 2D version of \cref{eq:gaussian}. For a certain pixel, the point-based rendering computes its color $\mathbf{C}$ by evaluating the blending of $N$ depth-ordered points overlapping that pixel:
\begin{equation}
    \mathbf{C} = \sum_{i=1}^{N} \mathbf{c}_i \alpha_i \prod_{j=1}^{i-1}(1-\alpha_j),
\label{eq:render}
\end{equation}
where $\mathbf{c}_i$ and $\alpha_i$ are the color and alpha value of Gaussian $i$, respectively.
The optimization of Gaussian parameters is then supervised by the reconstruction loss (difference between rendered and ground-truth images).

\subsection{Different Paradigms of Dynamic 3DGS}

The original 3DGS~\cite{3dgs} is optimized on a static scene and lacks the ability to model dynamic content.
Recently, researchers have made some promising attempts at dynamic 3DGS. These works can be abstracted into two main paradigms --- iterative and deformation-based dynamic modeling.

\parsection{Iterative 3DGS}
In this paradigm, some of the Gaussian parameters are iteratively updated in a frame-by-frame optimizing manner. The first frame is reconstructed with the original 3DGS for a static initialization. Then the centers and rotations of Gaussians are further tuned to adapt to the next frame, and so on. Other parameters and the number of Gaussians remain consistent over all timestamps. Once the entire process is complete, different spatial states of Gaussians across time form a coherent dynamic modeling.
This is an intuitive and highly interpretable extension of 3DGS. It can handle long-term dynamic content excellently. The main limitation is that those parts invisible in the initial frame can hardly be reconstructed well in later iterations. Besides, it requires a multi-camera setup since a monocular initialization for 3DGS remains a tough challenge.
A representative work using the iterative paradigm is D-3DGS~\cite{d3dgs}, on which our iteration-based framework is built.

\parsection{Deformation-based 3DGS}
Maintaining only a canonical representation and deforming 3D points based on time is a common approach for dynamic modeling in the NeRF research community~\cite{dnerf,nerfies}. Now with the explicit 3D Gaussian representation, more flexible choices of the deformation method are enabled, \eg, deformation field~\cite{4dgs,yang2023deformable} or time-dependent trajectory fitting~\cite{katsumata2023efficient}.
In this work, we take the first open-sourced deformation-based method 4D-GS~\cite{4dgs} as our baseline and build our deformation-based framework upon it. 4D-GS employs a deformation field to predict each Gaussian's offsets at a given timestamp compared with a mean canonical state. This deformation field can be decomposed into a multi-resolution HexPlane~\cite{hexplane} and an MLP-based decoder. Spatial and temporal adjacent Gaussians share similar deformation features by the design of HexPlane. For each Gaussian at a certain timestamp, the model queries the Hexplane with a 4D coordinate ($x\text{-}y\text{-}z\text{-}t$) and decodes its feature into the position, rotation, and scaling deformation values. 
This paradigm jointly optimizes the entire dynamic scene, enabling implicit global interactions of visual information. 
It is largely more time- and space-efficient and works on monocular videos, but may fail when faced with large movements.

It is worth noting that in this paper, we have \textit{no} intention of discussing which paradigm is more promising for dynamic 3DGS. Instead, we propose effective designs of motion-aware enhancement for both cases and hopefully advance the development of dynamic 3DGS.

\section{Method}

%

\begin{figure}[t]
  \centering
  \begin{minipage}[t]{1.0\linewidth}
    \includegraphics[width=1.0\linewidth]{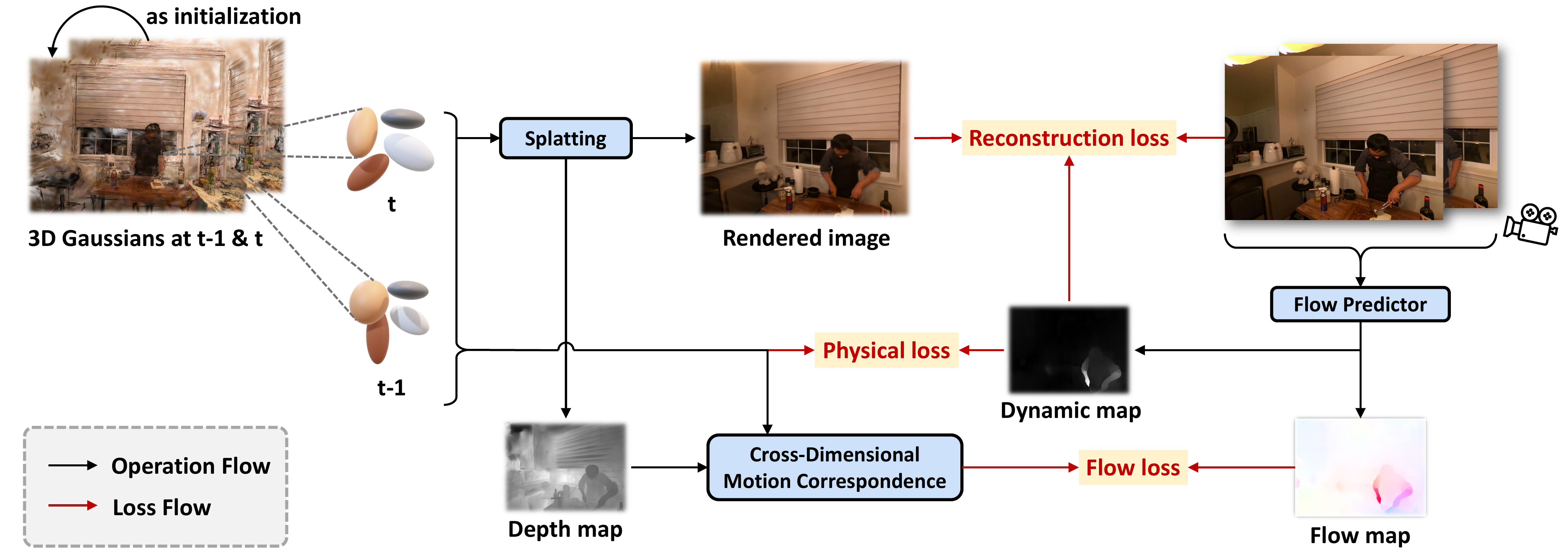}
    \dotfillin[1.0\linewidth]
  \end{minipage}
  \begin{minipage}[t]{1.0\linewidth}
    \vspace{0.7\baselineskip}
    \includegraphics[width=1.0\linewidth]{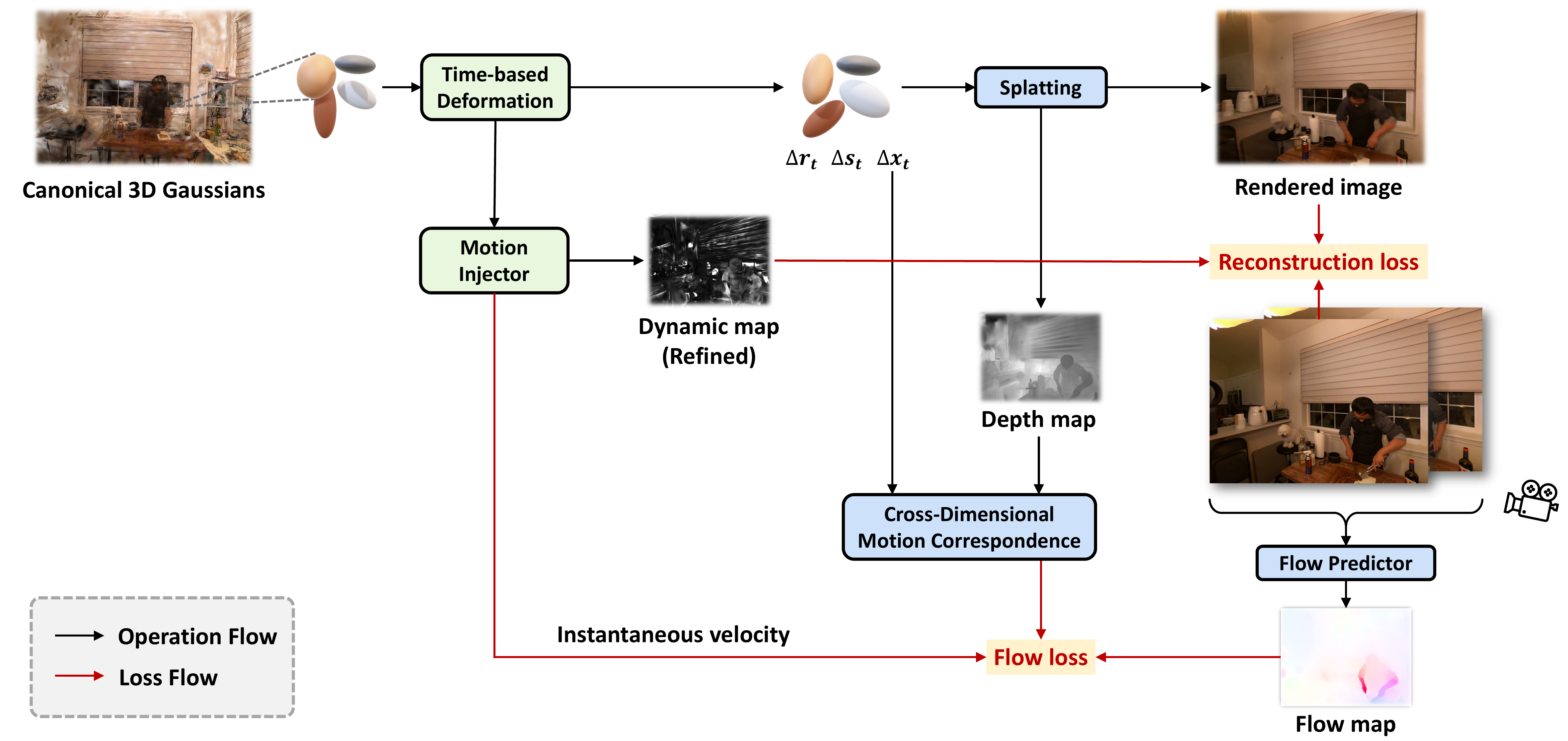}
  \end{minipage}
  \vspace{-3mm}
  \caption{\textbf{The proposed frameworks for iterative (above) and deformation-based (below) dynamic 3DGS.} We add motion-aware enhancement to both paradigms using the designed flow supervision and dynamic map. Moreover, for the deformation-based framework, a motion injector is further employed to handle the motion ambiguities by introducing auxiliary transient information to Gaussian features.
  }
  \label{fig:pipeline}
\end{figure}

For the dynamic reconstruction task, we propose a novel enhancement framework that fully exploits the motion cues from optical flow to improve existing dynamic 3DGS methods.
First, we establish a correspondence between 3D Gaussian movements and pixel-level flows (\cref{sec:motion-corr}).
Then we introduce our flow augmentation with additional insights into the uncertainty during training and collaboration with other supervisions (\cref{sec:flow}).
For the deformation-based paradigm that presents a harder optimization problem, an extra transient-aware deformation auxiliary module is proposed (\cref{sec:deform}).
Finally, we present the training process of our framework (\cref{sec:train}). The pipeline of our method is illustrated in \cref{fig:pipeline}.

\subsection{Cross-Dimensional Motion Correspondence}
\label{sec:motion-corr}

How to effectively model the temporal 3D scene flow is the core problem of dynamic scene reconstruction. 
In the context of 3DGS, such scene flow can be intuitively mapped to the movements of the Gaussian centers. 
To align those movements with 2D optical flow derived from image sequences, a dense cross-dimensional motion correspondence has to be established.

\begin{wrapfigure}{r}{0.5\linewidth}
\centering
\resizebox{1.0\linewidth}{!}{
\includegraphics{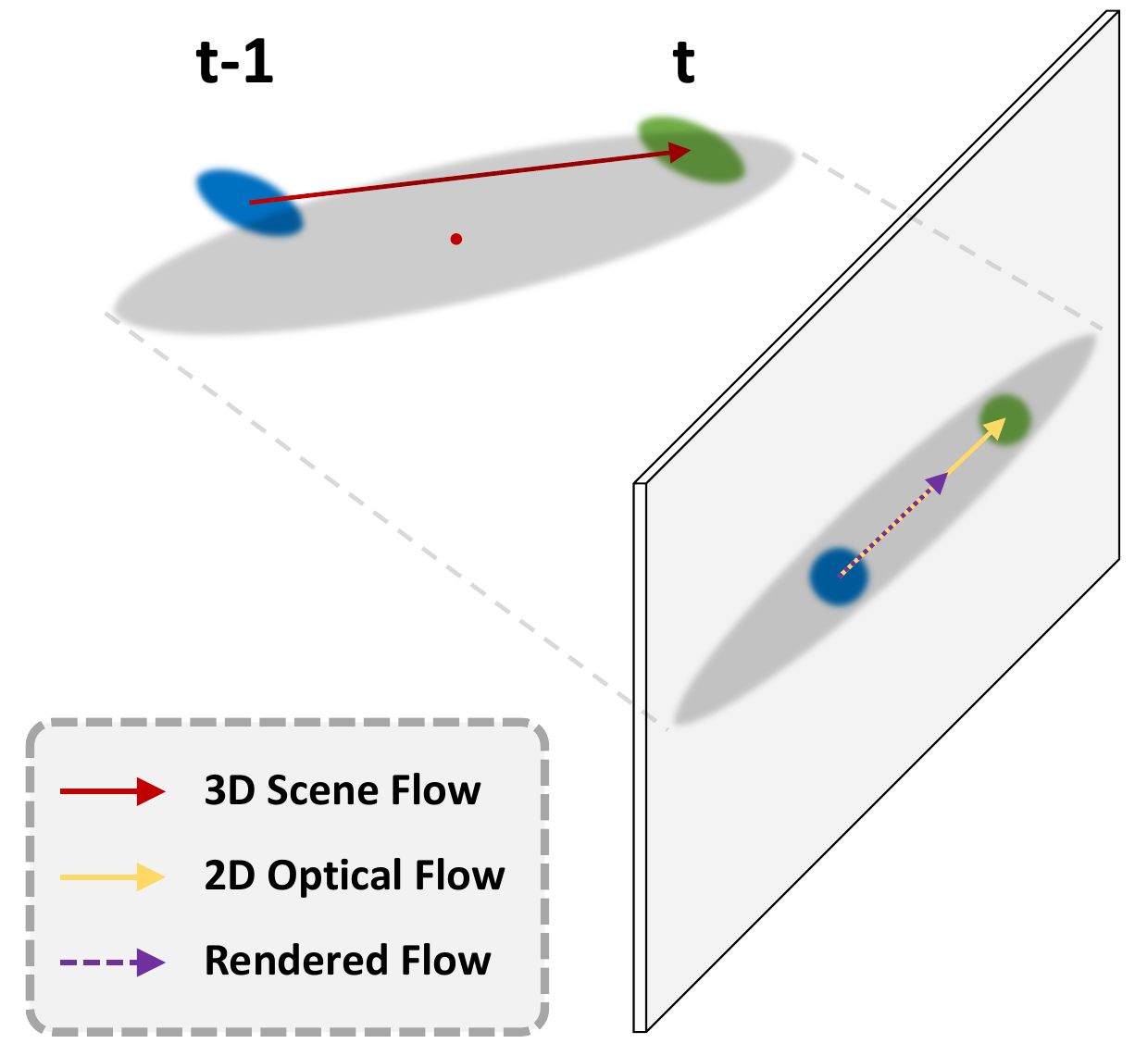}
}
\caption{\textbf{Illustration of the gap between rendered flow and actual optical flow.} Due to a static Gaussian in front, the 2D flow produced by the renderer is squeezed. Not only does this lead to supervision errors, but it also makes the front Gaussian incorrectly drift during optimization.}
\label{fig:render_flow}
\end{wrapfigure}

Considering the success of render-based depth supervision in NeRF~\cite{dsnerf,depthnerf,handnerf}, it seems a plausible solution to projecting the scene flow into the image plane using the differentiable renderer of 3DGS.
While the $\alpha$-blending in rendering can naturally deal with transparency and occlusion relationship, we found in practice that such render-based supervision presents poor performance when applied to optical flow, and sometimes is even prone to collapse of optimization.
The phenomenon can be explained from two views. 
On one side, unlike depth, optical flow is defined and predicted in 2D image plane, resulting in a non-negligible gap of render-based flow supervision (see \cref{fig:render_flow} for a more detailed illustration). 
On the other side, the long-distance (theoretically infinite) Gaussian influence makes the optimization vulnerable to noise and numerical instability. 
To address the above issues, we propose a more effective way to build the cross-dimensional bridge by performing foreground Gaussian searching and reprojecting.

\setlength{\intextsep}{0pt}

Specifically, when optimizing the motion for timestamp $t$, we start from the depth map of the scene produced as a byproduct of rendering at the last timestamp. 
For iterative 3DGS, the scene at $t-1$ is already successfully optimized. 
For deformation-based 3DGS, the scene is also generally well modeled after the initial coarse stage of training. 
Therefore, the depth map is moderately accurate in both paradigms. 
Then a pixel's 3D location $\mathbf{x}$ can be found via unprojection with the help of its depth value. 
We treat the $k$ nearest Gaussians around that 3D location as the foreground candidates for motion optimization. 
For a one-to-one correspondence, the Gaussian with the maximum contribution $\varphi_{k}(\mathbf{x})$ over this location (\cref{eq:gaussian}) can be selected. 
However, to make it more robust and efficient, we perform a ``soft select'' to use all $k$ nearby candidates along with their contributions in one iteration. 
All the 3D center $\mu$ of the candidates are reprojected into image plane at the current timestamp $t$ (camera view may change in a monocular scene), producing a set of 2D flows $\hat{\mathbf{F}}$ with gradients retained for motion optimization later: $\{\hat{\mathbf{F}}_{i,t}=\operatorname{Proj}(\mu_{i,t}-\mu_{i,t-1}) | i=1,2,\cdots,k\}$.

It is worth noting that the proposed method is \textit{not} a few-point approximation of the flow-rendering process mentioned earlier: all the foreground candidates of one pixel will be supervised by the same exact 2D optical flow.

\subsection{Flow Augmentation}
\label{sec:flow}

Once the projected scene flows for each pixel are obtained, we align them with the optical flow prior produced by a pretrained predictor~\cite{unimatch} from adjacent frames. Meanwhile, since the flow indicates the dynamic parts of a scene, it is also a useful tool to introduce dynamic awareness for other supervision signals.

\parsection{Flow loss with uncertainty}
Generally, an $L_1$ or $L_2$ loss between the projected and ground-truth flow can take effect as in previous works~\cite{nsff,chen2023bidirectional}. However, uncertainty in this process tends to hinder the optimization. On the one hand, flow predictor and camera calibration have their inherent errors, especially for monocular scenes with camera shake. On the other hand, the pixel-level flow is not always consistent with the motion of a Gaussian center, considering the existence of large Gaussians and complex movements involving scaling and rotation. To handle the above issues, we propose a new flow loss that takes uncertainty into consideration. Inspired by an effective solution in object detection~\cite{he2019bounding}, we use the KL-Divergence as the loss function for flow alignment:
\begin{equation}
    \begin{aligned}
    \mathcal{L}_{f} 
    & =\operatorname{KL}(P_t(f) \| \hat{P}_t(f)) \\
    & =\int P_t(f) \log P_t(f) \mathrm{d} f-\int P_t(f) \log \hat{P}_t(f) \mathrm{d} f \\
    & =\frac{(\mathbf{F}_t-\hat{\mathbf{F}}_t)^2}{2 \sigma_t^2}+\frac{\log \left(\sigma_t^2\right)}{2}+\frac{\log (2 \pi)}{2}-H\left(P_t(f)\right) \\
    & \propto \frac{(\mathbf{F}_t-\hat{\mathbf{F}}_t)^2}{2 \sigma_t^2}+\frac{1}{2} \log \left(\sigma_t^2\right).
    \end{aligned}
\end{equation}
It is intended to minimize the KL-Divergence between two distributions of flow $P_t(f)$ and $\hat{P}_t(f)$, which is finally derived to be a weighted difference between the flow prior $\mathbf{F}_t$ and the prediction $\hat{\mathbf{F}}_t$, along with a regularizer to avoid trivial solutions. We use the normalized Gaussian contribution as mentioned in \cref{eq:gaussian} in the formulation of the variance $\sigma_t^2$:
\begin{equation}
    1 / \sigma_t^2 = (\varphi_t / \max_k{\varphi_{k,t}}) \cdot c_t(\mathbf{d}),
\end{equation}
where $c_t(\mathbf{d})$ is the learnable view-dependent confidence. Overall, we expect that a Gaussian with less contribution and more uncertainty will produce a larger variance so that the flow loss term for that pixel can adaptively become lower.

\parsection{Loss with dynamic awareness}
Aside from direct motion supervision, we also leverage the flow prior to generate the ``dynamic map/mask'', thereby offering dynamic awareness to existing regularizations during optimization.

As the fundamental supervision, the pixel-level reconstruction (color) loss can be augmented with our dynamic map. We obtain the normalized magnitude of the 2D flow map and then weigh the prediction $\hat{\mathbf{C}}_t$ and ground truth $\mathbf{C}_t$ with it before calculating the $L_2$ image loss. Therefore, the flow prior plays the role of an attention map and guides the optimizer to attach more importance to regions with larger movements of the current frame. The refined color loss is given by
\begin{equation}
    \widetilde{\mathcal{L}}_{c}=
    (1-\lambda_c) \Vert\mathbf{C}_t - \hat{\mathbf{C}}_t\Vert_2^2 + \lambda_c \Vert\mathbf{C}_t \odot \mathbf{D}_t - \hat{\mathbf{C}}_t \odot \mathbf{D}_t\Vert_2^2,
\end{equation}
where $\mathbf{D}_t$ is the dynamic map, and $\odot$ means Hadamard product. We use a hyperparameter $\lambda_c$ to balance the dynamic-aware term and the original loss, in case the flow predictor fails to produce meaningful results for tiny motions.

Moreover, in the iterative D-3DGS~\cite{d3dgs}, a physical loss $\mathcal{L}_{p}$ is applied to a local region of Gaussians to encourage their motion similarity.
In order not to mix up the dynamic parts with the static background, a 3D foreground mask (jointly optimized using a segmentation loss) is adopted in $\mathcal{L}_{p}$. However, we argue that a constantly updating \textit{dynamic} mask is more suitable to distinguish between static and dynamic Gaussians, because the definition of the so-called ``foreground'' relies on 2D segmentation priors, and is usually quite different from the desired dynamic region. Therefore, we design a 3D dynamic mask based on the dynamic map $\mathbf{D}_t$. Using the process described in \cref{sec:motion-corr}, it is easy to find a dense correspondence for the pixel-level motion magnitude. We combine the flow information from all viewpoints of the current frame and set a threshold to get the dynamic labels for all Gaussians. In this way, we upgrade $\mathcal{L}_{p}$ to a more accurate dynamic-aware version with no extra cost like the segmentation.

\subsection{Transient-aware Deformation Auxiliary}
\label{sec:deform}

Compared with iterative 3DGS, the deformation-based paradigm tries to solve a more difficult optimization problem where scenes at all timestamps are jointly modeled. The aforementioned flow augmentation can still be applied to it, but the performance gain is somehow minor. We blame this on its different modeling process. In the iterative paradigm, the well-optimized scene at $t-1$ can be viewed as a fixed anchor, with which the flow priors guide Gaussians to a clear destination at the current time $t$. However, in deformation-based 3DGS, neither scenes at $t-1$ and $t$ are fixed. Without the anchor, simply optimizing the relative motion of the two has ambiguities and can hardly achieve an ideal effect. Therefore, auxiliary designs are essential to enhance the deformation.

\parsection{Motion injector}
Considering the HexPlane structure of our baseline~\cite{4dgs} for deformation-based 3DGS, we propose to explicitly inject transient information into the time-variant voxel features. A velocity field $\mathbf{\Theta}_v$ is employed to decode the position- and time-dependent feature $\mathbf{g}_t$ of each Gaussian into a 3D instantaneous velocity $\mathbf{v}_t$. Through dense correspondence searching (\cref{sec:motion-corr}), $\mathbf{v}_t$ is aligned with optical flow $\mathbf{f}_t$ divided by the time interval $\mathrm{\Delta} t$ (only if it belongs to a visible foreground Gaussian from the current view). To exploit time-contextual cues, the velocity field also takes features from adjacent frames as inputs and integrates them with the current frame.
This alignment term is added to the flow loss proposed in \cref{sec:flow} and rewritten as follows:
\begin{align}
    \widetilde{\mathcal{L}}_{f} &= \mathcal{L}_{f} + \sum\nolimits_{i} \mathbbm{1}(i) \Vert \operatorname{Proj}(\mathbf{v}_{i,t}) \mathrm{\Delta} t - \mathbf{f}_{i,t} \Vert_1, \\
    \mathbf{v}_{i,t} &= \mathbf{\Theta}_v(\mathbf{g}_{i,t},\mathbf{g}_{i,t-1},\mathbf{g}_{i,t+1}),
\end{align}
where $\operatorname{Proj}(\cdot)$ is the reprojecting operation, and $\mathbbm{1}(\cdot)$ is an indicator function: $\mathbbm{1}(i)=1$ iff Gaussian $i$ is selected via foreground searching from current view (\cref{sec:motion-corr}). With $\widetilde{\mathcal{L}}_{f}$, feature optimization no longer relies solely on the constraints over relative offsets of Gaussian centers across time. Transient motion information without ambiguity is now directly injected into the current timestamp.\footnote{Motion here should not be confused with the position deformation in the original paradigm. The former indicates instantaneous movements between adjacent frames, while the latter is based on a canonical scene with time accumulation.}

\parsection{Dynamic map refinement}
In \cref{sec:flow}, we propose to use the normalized flow magnitude as a dynamic map to craft dynamic-aware reconstruction loss. Now with our motion injector, the predicted instantaneous velocity can be a perfect replacement for the ground-truth optical flow. The reason is twofold. First, velocity comes from a learnable network, which can cover some errors of the prior flow prediction and produce a refined dynamic map. Second, the deformation-based paradigm enables the modeling of monocular scenes, where optical flows between adjacent frames are affected by camera displacements. In this case, all pixels in an image will be marked as dynamic, but we only care about the motion of the scene itself. The refined dynamic map can naturally solve this problem by projecting 3D velocity into a common camera plane.

\subsection{Optimization}
\label{sec:train}

Our motion-aware dynamic 3DGS framework is optimized in an end-to-end manner. The overall training loss for the iterative paradigm is given by
\begin{equation}
    \mathcal{L}^{I}=\widetilde{\mathcal{L}}_c + \lambda_{p} \widetilde{\mathcal{L}}_{p} + \lambda_{f} \mathcal{L}_{f},
\end{equation}
where the reconstruction loss $\widetilde{\mathcal{L}}_c$ and physical loss $\widetilde{\mathcal{L}}_{p}$ are both equipped with dynamic awareness. Note that the segmentation loss~\cite{d3dgs} is not included since the flow-based dynamic map is more competent than foreground segmentation.
For our deformation-based framework, the transient-aware auxiliary introduces an upgraded version of flow loss $\widetilde{\mathcal{L}}_{f}$. Given that the adopted deformation field based on HexPlane has already ensured the similarity between neighboring Gaussians, there is no need for the physically-based constraints anymore. Thus, the refined loss for the deformation-based paradigm is even more concise:
\begin{equation}
    \mathcal{L}^{D}=\widetilde{\mathcal{L}}_c + \lambda_{f} \widetilde{\mathcal{L}}_{f}.
\end{equation}
Moreover, we adjust $\lambda_{f}$ with a schedule similar to warmup and cosine annealing. Therefore, the flow loss takes into effect after a coarse geometry (depth) is prepared. As the motion modeling approaches perfection, $\lambda_{f}$ progressively decays to make the network focus more on textual details.

\section{Experiments}
\label{sec:exp}

\subsection{Experimental Settings}

Our method is evaluated on three datasets with different setups.
1) PanopticSports~\cite{d3dgs}, which contains sequences of human motions and object interactions (lab-collected multi-view data).
2) Neural 3D Video dataset~\cite{dynerf} (Neu3DV), which includes high-resolution videos captured with synchronized fixed cameras (in-the-wild multi-view data). We follow \cite{4dgs} to evaluate the methods exclusively on \textit{cook\_spinach}, \textit{cut\_roasted\_beef}, and \textit{sear\_steak}.
3) HyperNeRF dataset~\cite{hypernerf}, which comprises videos ranging from 8 to 15 seconds captured by smartphones (in-the-wild monocular data). We follow \cite{4dgs} to evaluate the methods exclusively on \textit{chickchicken}, \textit{split-cookie}, \textit{cut-lemon1}, and \textit{vrig-3dprinter}.
Note that the synthetic D-NeRF dataset~\cite{dnerf} is not included, since its camera setup is designed to mimic a monocular setting by randomly teleporting between adjacent timestamps. Such dramatic changes in viewpoints will hinder the effectiveness of our image-based flow augmentation.
%
%
Following previous works, we evaluate the reconstruction with PSNR, SSIM, and LPIPS\footnote{For comparison on Neu3DV (\cref{tab:dynerf}), we report AlexNet-based LPIPS for consistency with previously reported benchmarks, while other results still use VGG-based metric.}~\cite{lpips}.

\subsection{Comparison Results}

\begin{figure}[t]
  \centering
  \includegraphics[width=\linewidth]{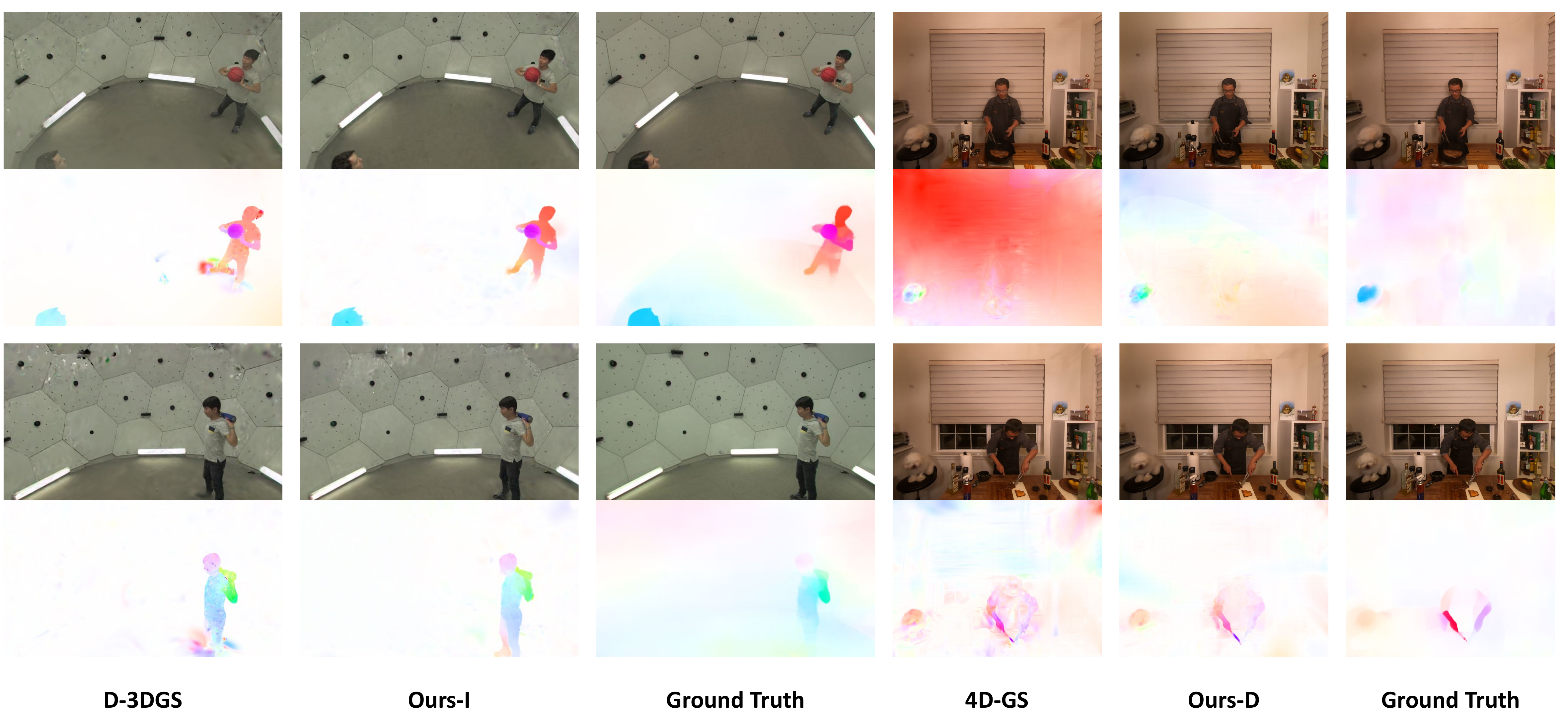}
  \caption{\textbf{Qualitative comparison on PanopticSports~\cite{d3dgs} (left) and Neu3DV~\cite{dynerf} (right).} The dense optical flow maps below the images are produced by rendering the Gaussian movements (according to the Middlebury color coding~\cite{baker2011database}). Those render-based flows serve as a useful tool for qualitative visualization, despite being a poor choice of supervision signal.
  }
  \label{fig:sports_dynerf}
\end{figure}

\begin{table}[t]
  \centering
  \caption{\textbf{Quantitative comparison on PanopticSports~\cite{d3dgs}.}
  \textbf{Bold} denotes the best performance, and \underline{underline} denotes the second place.
  ``Ours-I'' here refers to the proposed iterative framework based on D-3DGS~\cite{d3dgs}.
  }
    \begin{tabular}{c|l|ccccccc}
    \toprule
    Metrics & Method & Juggle & Boxes & Softball & Tennis & Football & Basketball & Mean \\
    \midrule
    \multirow{3}[0]{*}{PSNR $\uparrow$}
          & 3DGS-O~\cite{d3dgs} & 28.19  & 28.74  & \underline{28.77}  & 28.03  & \underline{28.49}  & 27.02  & 28.21  \\
          & D-3DGS~\cite{d3dgs} & \underline{29.48}  & \underline{29.46}  & 28.43  & \underline{28.11}  & \underline{28.49}  & \underline{28.22}  & \underline{28.70}  \\
          & Ours-I  & \textbf{29.55} & \textbf{29.60} & \textbf{29.54} & \textbf{28.19} & \textbf{29.71} & \textbf{29.60} & \textbf{29.37} \\
    \midrule
    \multirow{3}[0]{*}{SSIM $\uparrow$}
          & 3DGS-O~\cite{d3dgs} & 0.91  & 0.91  & \underline{0.91}  & 0.90  & 0.90  & 0.89  & 0.90  \\
          & D-3DGS~\cite{d3dgs} & \textbf{0.92} & \textbf{0.91} & \underline{0.91}  & \textbf{0.91} & \underline{0.91}  & \underline{0.91}  & \underline{0.91} \\
          & Ours-I  & \textbf{0.92} & \textbf{0.91} & \textbf{0.92} & \textbf{0.91} & \textbf{0.92} & \textbf{0.92} & \textbf{0.92} \\
    \midrule
    \multirow{3}[0]{*}{LPIPS $\downarrow$}
          & 3DGS-O~\cite{d3dgs} & \underline{0.15} & \textbf{0.15} & \textbf{0.14} & \textbf{0.16} & \textbf{0.16} & \underline{0.18}  & \textbf{0.16} \\
          & D-3DGS~\cite{d3dgs} & \underline{0.15} & 0.17  & 0.19  & 0.17  & 0.19  & \underline{0.18}  & 0.17  \\
          & Ours-I  & \textbf{0.16} & \underline{0.16}  & \underline{0.16}  & \textbf{0.16} & \textbf{0.16} & \textbf{0.16} & \textbf{0.16} \\
    \bottomrule
    \end{tabular}%
  \label{tab:sports}%
\end{table}%

\begin{table}[t]
  \centering
  \caption{\textbf{Quantitative comparison on Neu3DV~\cite{dynerf}.}
  \textbf{Bold} denotes the best performance, and \underline{underline} denotes the second place.
  ``Ours-I'' and ``Ours-D'' refer to our iterative and deformation-based framework, respectively.
  }
    \resizebox{1.0\linewidth}{!}{
    \begin{tabular}{l|ccc|ccc|ccc|ccc}
    \toprule
    \multicolumn{1}{c|}{Scene} & \multicolumn{3}{c|}{\textit{cook\_spinach}} & \multicolumn{3}{c|}{\textit{cut\_roasted\_beef}} & \multicolumn{3}{c|}{\textit{sear\_steak}} & \multicolumn{3}{c}{Mean} \\
    \midrule
    \multicolumn{1}{c|}{Metrics} & PSNR & SSIM & \multicolumn{1}{c|}{LPIPS} & PSNR & SSIM & \multicolumn{1}{c|}{LPIPS} & PSNR & SSIM & \multicolumn{1}{c|}{LPIPS} & PSNR & SSIM & LPIPS \\
    \midrule
    NeRFPlayer~\cite{song2023nerfplayer} & 30.56  & 0.9290  & 0.1130  & 29.35  & 0.9080  & 0.1440  & 29.13  & 0.9080  & 0.1380  & 29.68  & 0.9150  & 0.1317  \\
    HyperReel~\cite{attal2023hyperreel} & 32.30  & 0.9410  & 0.0890  & \textbf{32.92}  & \underline{0.9450}  & 0.0840  & 32.57  & 0.9520  & 0.0770  & \underline{32.60}  & 0.9460  & 0.0833  \\
    D-3DGS~\cite{d3dgs} & \underline{32.97}  & \underline{0.9474}  & 0.0870  & 30.72  & 0.9410  & 0.0900  & \underline{33.68}  & \underline{0.9552}  & 0.0790  & 32.46  & \underline{0.9479}  & 0.0853  \\
    4D-GS~\cite{4dgs} & 31.98  & 0.9385  & \underline{0.0564}  & 31.56  & 0.9394  & \underline{0.0619}  & 31.20  & 0.9486  & \underline{0.0455}  & 31.58  & 0.9422  & \underline{0.0546}  \\
    \midrule
    Ours-I & \textbf{33.26} & \textbf{0.9536} & 0.0822 & \underline{32.64} & \textbf{0.9462} & 0.0820 & \textbf{33.92} & \textbf{0.9570} & 0.0754 & \textbf{33.27} & \textbf{0.9523} & 0.0799 \\
    Ours-D & 32.10 & 0.9367 & \textbf{0.0559} & 32.56 & 0.9414 & \textbf{0.0589} & 31.60 & 0.9508 & \textbf{0.0452} & 32.09 & 0.9430 & \textbf{0.0533} \\
    \bottomrule
    \end{tabular}%
    }
  \label{tab:dynerf}%
\end{table}%

\cref{tab:sports} summarizes the performance of our iterative framework and two baselines: D-3DGS~\cite{d3dgs} and an online-iterative version of the original 3DGS (3DGS-O)~\cite{d3dgs}.
\cref{tab:dynerf} presents the comparison between our frameworks and three other competitors, including 4D-GS~\cite{4dgs} that we build our deformation-based framework upon.
On these two multi-view benchmarks, our method shows significant advantage over the competitors in most scenes. Both paradigms equipped with our motion-aware enhancement have shown noteworthy improvements. Notably, in \cref{tab:dynerf}, previous 3DGS works are generally inferior to the latest NeRF-based method~\cite{attal2023hyperreel}. Once enhanced with our designs, their performance becomes more competitive, and the iterative framework even outperforms it by a large margin.
The qualitative results on multi-view datasets are exhibited in \cref{fig:sports_dynerf}. Our method boosts reconstruction performance by effectively introducing motion information into the paradigms. We can observe much less background noise and more precise motion in the rendered flow map. Notably, with our uncertainty-aware design, the model is not misled by intrinsic errors of the ground-truth flow as it adaptively produces more reasonable motions (\eg, left part of \cref{fig:sports_dynerf}).

\begin{figure}[t]
  \centering
  \begin{minipage}[t]{0.48\linewidth}
    \centering
    \includegraphics[width=\linewidth]{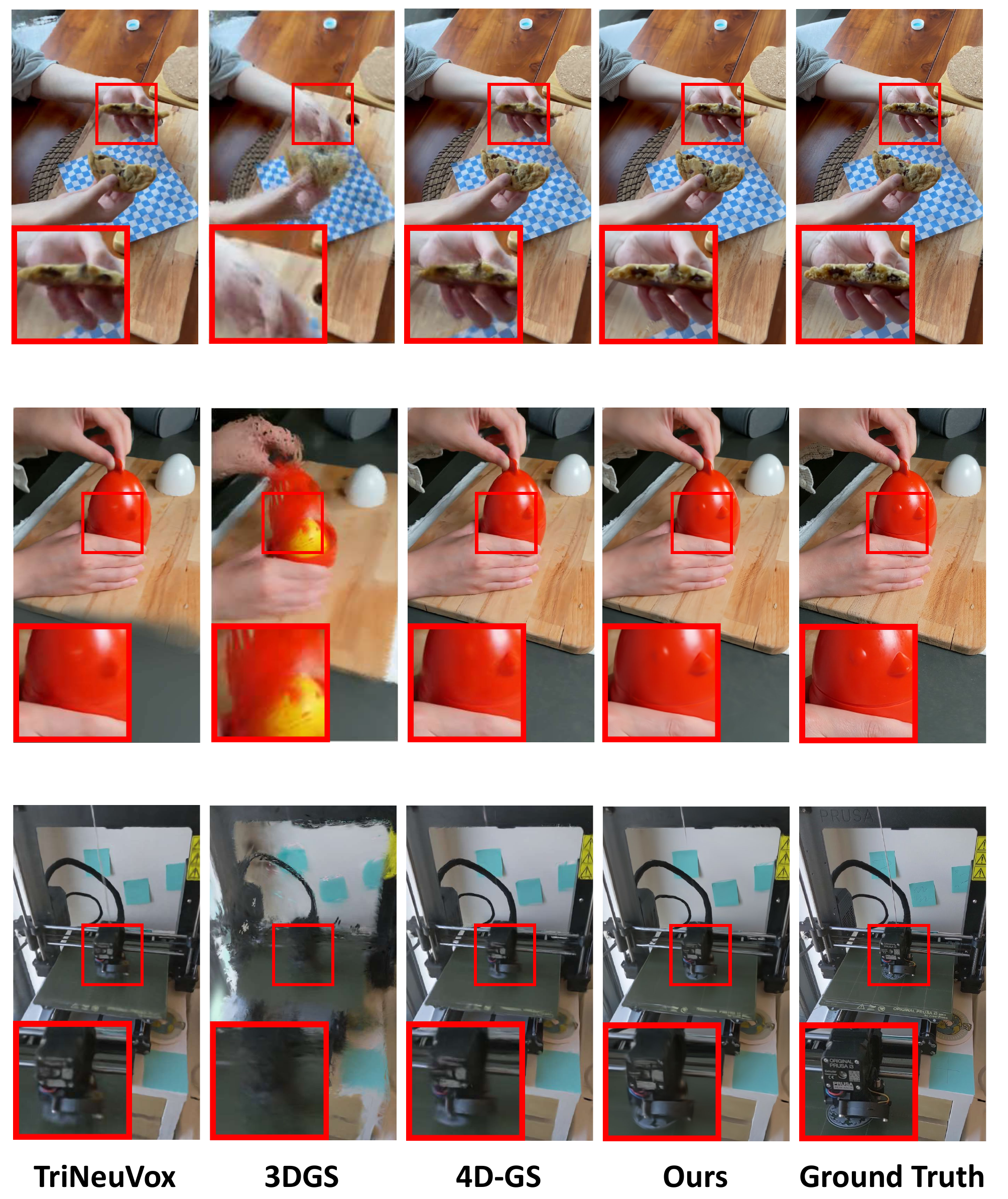}
    \caption{
    \textbf{Qualitative comparison on the monocular in-the-wild scenes of HyperNeRF~\cite{hypernerf} dataset.}
    We show zoomed-in details of the dynamic region for each result.
    }
    \label{fig:hypernerf}
  \end{minipage}%
  \quad
  \begin{minipage}[t]{0.48\linewidth}
    \centering
    \includegraphics[width=\linewidth]{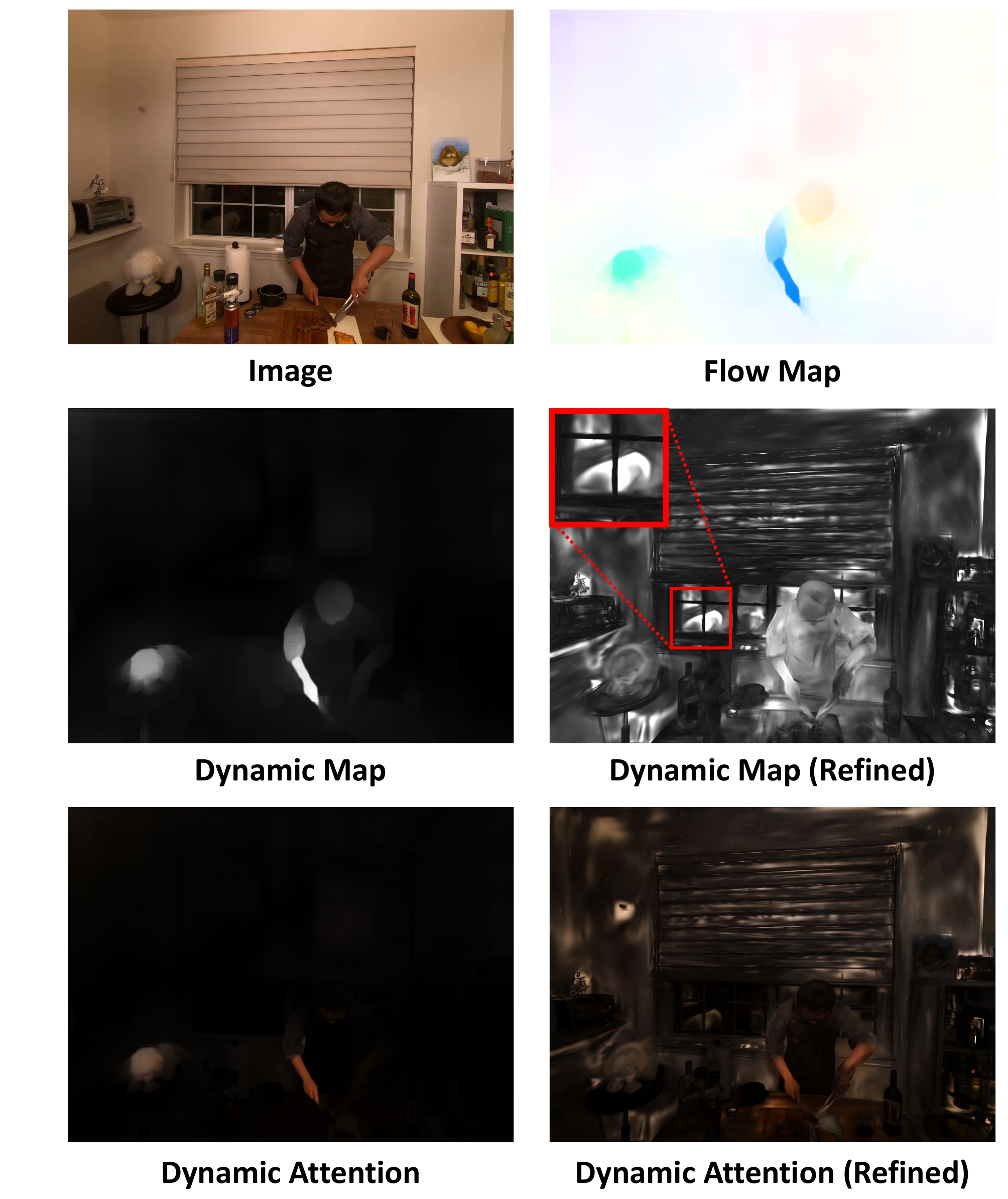}
    \caption{\textbf{Visualization of the flow map, dynamic maps, and their attention regions.}
    The coarse dynamic map is derived directly from the flow map and shows a quite limited region of interest.
    }
    \label{fig:dynamic_map}
  \end{minipage}
\end{figure}

Monocular scenes serve as an even better testbed for our framework. Without flow priors, some motion cues can still be naturally extracted from sufficient viewpoints, but generally fail to be noticed under a monocular setting, where the camera also has displacements.
As shown in \cref{tab:hypernerf}, our deformation-based framework outperforms all the baselines across all the metrics.
In \cref{fig:hypernerf}, we can clearly observe the superiority of our method. The motion parts are modeled well with less blur, while the static regions remain clear and detailed.

\begin{table}[t]
  \centering
  \begin{minipage}[t]{0.485\linewidth}
  \centering
  \caption{\textbf{Quantitative comparison on the monocular HyperNeRF dataset~\cite{hypernerf}.}
  \textbf{Bold} denotes the best performance, and \underline{underline} denotes the second place.
  ``Ours-D'' here refers to the proposed deformation-based framework built upon 4D-GS~\cite{4dgs}.
  }
  \resizebox{1.0\linewidth}{!}{
    \begin{tabular}{l|ccc}
    \toprule
    Method & PSNR $\uparrow$ & SSIM $\uparrow$ & LPIPS $\downarrow$ \\
    \midrule
    TiNeuVox-B\cite{fang2022fast} & 26.87  & 0.75  & 0.37  \\
    3DGS~\cite{3dgs}  & 20.84  & 0.70  & 0.45  \\
    4D-GS~\cite{4dgs} & \underline{26.98}  & \underline{0.78}  & \underline{0.31}  \\
    \midrule
    Ours-D & \textbf{27.87} & \textbf{0.80} & \textbf{0.27} \\
    \bottomrule
    \end{tabular}%
    }
  \label{tab:hypernerf}%
  \end{minipage}%
  \quad
  \begin{minipage}[t]{0.46\linewidth}
  \centering
  \caption{\textbf{Effect of sparser viewpoints for multi-view scenes.}
  For the iterative paradigm (upper), we cut the available views from 27 to 10 (except the first frame).
  For the deformation-based paradigm (lower), we reduce the number of views by 5 for all frames.
  }
  \resizebox{1.0\linewidth}{!}{
    \begin{tabular}{l|ccc}
    \toprule
    Method & PSNR $\uparrow$ & SSIM $\uparrow$ & LPIPS $\downarrow$ \\
    \midrule
    D-3DGS~\cite{d3dgs} & 24.3256 & 0.8254 & 0.2735 \\
    Ours-I  & \textbf{25.7227} & \textbf{0.8599} & \textbf{0.2223} \\
    \midrule
    4D-GS~\cite{4dgs} & 30.5072 & 0.9309 & 0.1639 \\
    Ours-D  & \textbf{31.5064} & \textbf{0.9373} & \textbf{0.1634} \\
    \bottomrule
    \end{tabular}%
    }
  \label{tab:sparser}%
  \end{minipage}%
\end{table}%

\subsection{Study on the Efficiency of Dynamic Reconstruction}

The proposed method enables efficient dynamic reconstruction for three reasons.
1) With modern optical flow predictors, the prior acquisition can be done in real time. Since our flow augmentation only affects the optimizing pipeline, no extra cost is introduced in rendering.
2) Motion regularization helps produce modeling results with less Gaussian and motion redundancy, especially for monocular reconstruction.
3) Our method still achieves competitive performance with sparser viewpoints for multi-view scenes.
We will further interpret the last two points by providing more experimental results.

\begin{figure}[!t]
  \centering
  \includegraphics[width=0.95\linewidth]{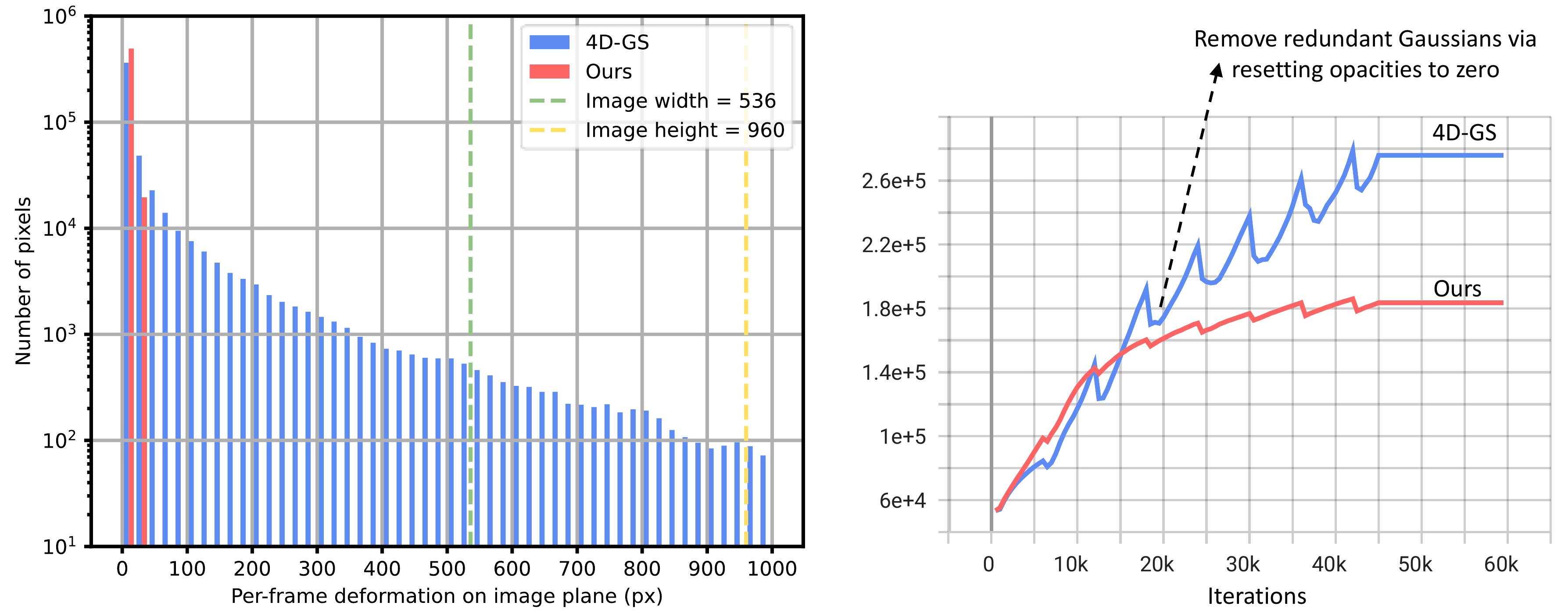}
  \caption{\textbf{Left: histogram of the Gaussian deformation distribution between adjacent frames.} The baseline 4D-GS tends to produce unreasonably large movements. Many foreground Gaussians even escape from the image plane.
  \textbf{Right: the growing number of Gaussians as the training progresses.} Our framework enables more efficient dynamic modeling by naturally restraining the growth of Gaussians.
  Both graphs are plotted with the results on the \textit{chickchicken} scene of HyperNeRF dataset.
  }
  \label{fig:efficient}
\end{figure}

\parsection{Reducing redundancy for monocular scenes}
When digging into the reconstruction results of monocular scenes, we find that the baseline result has redundancy when modeling dynamic parts. The model is prone to a local optimum where many Gaussians are kept outside of the current view and then moved into the image plane at a certain timestamp. It can be validated by \cref{fig:efficient} (left) that the foreground Gaussian movements are unreasonably large compared with the image size. In this way, much more Gausssians are needed to force a per-frame dynamic fitting. However, this approach can only model coarse object motion and significantly degrades the rendering quality.
For comparison, with our motion-aware guidance, the model learns to reuse existing Gaussians as much as possible, which presents a better dynamic modeling with cross-time correspondence. As illustrated in \cref{fig:efficient} (right), our method can achieve better performance and less overhead without any additional design for Gaussian pruning. Please refer to the supplementary material for a detailed per-sequence comparison of the space and time efficiency in monocular scenes.

\parsection{Robustness under sparser viewpoints for multi-view scenes}
We try to use sparser viewpoints for dynamic modeling in two multi-view datasets.
As shown in \cref{tab:sparser}, while the baselines suffer from significant degradation, our method can still achieve competitive performance. This reveals the potential of our motion-aware design for sparse-view 3DGS reconstruction.

\subsection{Ablation Study}

In \cref{tab:ablation}, we conduct ablative experiments on \textit{chickchicken} of HyperNeRF dataset to validate the effectiveness of essential components in our work.

\parsection{Motion correspondence and flow supervision}
We compare the proposed dense correspondence searching and KL-based flow loss with the flow rendering and a vanilla $L_1$ loss. The render-based supervision has defects as we discussed in \cref{sec:motion-corr} and leads to performance degradation. Meanwhile, without uncertainty handling, the model is easily affected by errors of priors.

\parsection{Deformation auxiliary}
For the deformation-based paradigm, our motion injector is vital, without which the model is solving an ambiguous problem and shows limited performance improvement solely with flow supervision.

\parsection{Dynamic awareness}
It can be observed that the dynamic awareness we introduce to the reconstruction loss does guide the model to focus more on motion parts.
Then our refined dynamic map further gives attention to regions needed for accuracy and covers errors in the flow prior. As demonstrated in \cref{fig:dynamic_map}, the raw flow map fails to extract motion for the reflection of the dog in the window, while our refined dynamic map keenly notices and guides the model to enhance the scene representation there.

\begin{table}[t]
  \centering
  \caption{\textbf{Ablations on the components of our proposed framework.}
  }
  \setlength{\tabcolsep}{1em}{
    \begin{tabular}{l|ccc}
    \toprule
          & PSNR $\uparrow$ & SSIM $\uparrow$ & LPIPS $\downarrow$ \\
    \midrule
    Baseline (4D-GS~\cite{4dgs}) & 25.9856  & 0.7566  & 0.4434  \\
    render-based flow loss & 25.8640  & 0.7487  & 0.4552  \\
    $L_1$ flow loss & 26.4666  & 0.7904  & 0.3554  \\
    w/o motion injector & 26.0628  & 0.7749  & 0.4147  \\
    w/o dynamic map & 26.5940  & 0.7915  & 0.3479  \\
    dynamic map w/o refinement & 26.8047  & 0.7977  & 0.3243  \\
    Ours  & \textbf{26.9181} & \textbf{0.8004} & \textbf{0.3170} \\
    \bottomrule
    \end{tabular}%
    }
  \label{tab:ablation}%
\end{table}%

\section{Conclusion and Discussion}

In this paper, we propose the motion-aware 3DGS, a novel enhancement framework for efficient dynamic scene reconstruction.
How to leverage motion cues from optical flow to enhance 3DGS-based modeling is a non-trivial challenge. By developing several elaborate strategies including uncertainty-aware flow augmentation and transient-aware deformation auxiliary, we provide an effective solution for both iterative and deformation-based paradigms of dynamic 3DGS.
Comprehensive experiments on prevalent datasets demonstrate the superiority of our approach and its huge potential for future investigation.

Despite the merits, there is still room for improvement in this field.
First, since our work relies on optical flow predictions from 2D images, motion blur is an intractable obstacle in dynamic modeling. Sometimes the model is actually overfitting the blur parts and hindering the temporal consistency of Gaussians.
Second, although specific designs are adopted to mitigate the effect of errors in flow priors, it is still non-trivial to break through the limitation of motion uncertainty, especially for monocular scenes. Exploring more useful priors could be a promising avenue for future work.

%
%
\bibliographystyle{splncs04}
\bibliography{egbib}
\end{document}